\documentclass[letterpaper]{article} 
\usepackage{aaai25}  
\usepackage{times}  
\usepackage{helvet}  
\usepackage{courier}  
\usepackage[hyphens]{url}  
\usepackage{graphicx} 
\usepackage{natbib}  
\usepackage{caption} 
\usepackage{times}  

\usepackage[table]{xcolor}
\usepackage{helvet}  
\usepackage{courier}  
\usepackage{graphicx} 
\usepackage{natbib}  
\usepackage{caption}
\usepackage{algorithm}
\usepackage{algorithmic}
\usepackage{amsmath}
\usepackage{multirow}
\usepackage{bm}
\usepackage{booktabs}
\usepackage{amsfonts}
\usepackage{newfloat}
\usepackage{listings}
\usepackage{breakurl}
\usepackage{newfloat}
\usepackage{listings}
\DeclareCaptionStyle{ruled}{labelfont=normalfont,labelsep=colon,strut=off} 
\floatstyle{ruled}
\newfloat{listing}{tb}{lst}{}
\floatname{listing}{Listing}
\title{Retrieval-Augmented Visual Question Answering\\ via Built-in Autoregressive Search Engines}
\author{
    Xinwei Long\textsuperscript{1},
    Zhiyuan Ma\textsuperscript{1},
    Ermo Hua\textsuperscript{1},
    Kaiyan Zhang\textsuperscript{1},
    Biqing Qi\textsuperscript{2},
    Bowen Zhou\textsuperscript{1}\thanks{Corresponding author.}
}
\affiliations{
    \textsuperscript{\rm 1}Department of Electronic Engineering, Tsinghua University\\
    \textsuperscript{\rm 2}Shanghai Artificial Intelligence Laboratory\\

    longxw22@mails.tsinghua.edu.cn
}

\usepackage{bibentry}

\begin{document}

\maketitle

\begin{abstract}
Retrieval-augmented generation (RAG) has emerged to address the knowledge-intensive visual question answering (VQA) task.
Current methods mainly employ separate retrieval and generation modules to acquire external knowledge and generate answers, 
respectively.
We propose ReAuSE, an alternative to the previous RAG model for the knowledge-based VQA task, which seamlessly integrates knowledge retriever into the generative multi-modal large language model, serving as a built-in search engine.
Specifically, our model functions both as a generative retriever and an accurate answer generator. It not only helps retrieve documents from the knowledge base by producing identifiers for each document, but it also answers visual questions based on the retrieved documents.
Furthermore, we also propose a reinforced retrieval calibration module from relevance feedback to improve retrieval performance and align with the preferences for accurate answer generation.
Extensive experiments on two representative OKVQA and A-OKVQA datasets demonstrate significant improvements ranging from 2.9\% to 9.6\% across all evaluation metrics when compared to strong baselines. 

\end{abstract}

%

\section{Introduction}
The Visual Question Answering (VQA) task aims to answer questions based on a user-provided image, which has received significant attention from CV and NLP community~\cite{Antol_2015_ICCV,hu2017learning,shen2023git,sunprogram,DBLP:conf/naacl/ZhuQZLLZ24}.
Early VQA methods~\cite{mascharka2018transparency,gao2019dynamic} mainly focus on understanding visual elements within the image.
Recently, the research trend of VQA has shifted towards knowledge-intensive scenarios~\cite{Shah_Mishra_Yadati_Talukdar_2019}, requiring the incorporation of external knowledge and joint reasoning over multi-modal content to generate accurate answers.
However, existing methods generally face challenges in effectively acquiring relevant information from large-scale knowledge bases using multi-modal queries~\cite{DBLP:conf/nips/LinCMCB23}.

\begin{figure}[t]
\centering
\includegraphics[width=1.0\columnwidth]{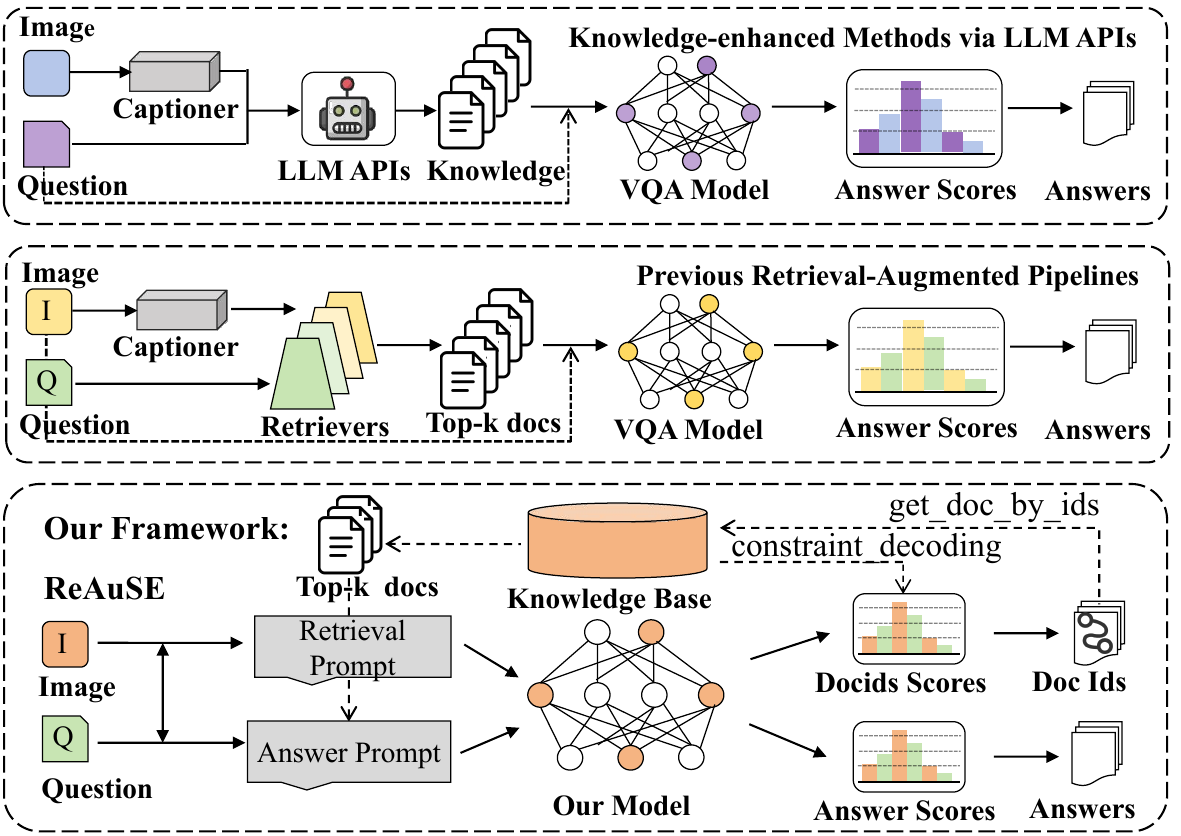} 
\caption{Comparing with the Paradigm of Previous Knowledge-Based VQA Methods.}
\label{fig1}
\end{figure}

Retrieval-augmented generation (RAG)~\cite{chan2024rq,chen2024benchmarking} has recently emerged as a promising approach for knowledge-based visual question answering (KBVQA) tasks~\cite{DBLP:conf/cvpr/0013PTRWN22,DBLP:conf/emnlp/LinB22,DBLP:conf/emnlp/ChenHCVC22}.
RAG-based approaches typically consist of two separate stages: retrieval and generation. 
In the first retrieval stage, these methods usually integrate multiple discriminative retrievers, each designed for specific purposes such as image-to-text or text-to-text retrieval.
Afterward, in the second answer generation stage, these methods typically use generative multi-modal large language models (MLLM) to produce the final result.
Despite achieving success in some benchmarks~\cite{DBLP:conf/cvpr/MarinoRFM19,DBLP:conf/eccv/SchwenkKCMM22}, this workflow still encounters several limitations.
1) Current methods sequentially invoke models in the pipeline for feature engineering, retrieval, and answer generation, requiring the integration of multiple heterogeneous models.
2) Moreover, these methods typically combine generative answer generators with discriminative retrievers. The disparate model architectures make it challenging for retrievers to further optimize their performance based on the feedback from the answer generator.
Consequently, the research question arises: \textbf{\textit{how can we integrate knowledge retrieval and answer generation into a homogeneous generative model?}}


To address the above issue, we propose ReAuSE, a novel \underline{Re}trieval-augmented framework with built-in \underline{Au}toregressive \underline{S}earch \underline{E}ngines for knowledge-based VQA tasks, which seamlessly integrates knowledge retrieval into the generative MLLM.
ReAuSE takes advantage of the fact that MLLMs can serve as virtual knowledge warehouses~\cite{pan2024unifying}, 
recognizing the documents that a multi-modal query can be linked to.
Therefore, ReAuSE abandons the discriminative retrieval paradigm that computing the similarity between the query and document one by one,
whereas directly generates the document identifier in an autoregressive manner, where each identifier corresponds to a document within the knowledge base.
We define the document identifiers as a sequence of tokens that appears at least once within a document in the knowledge base, thus enabling effective and efficient mapping to the document.
Subsequently, we propose a reinforced retrieval calibration method based on relevance feedback to further enhance retrieval performance.
To collect relevance preference data, we employ a MLLM as a reward model, which inputs sampled documents and questions into this model and assesses document relevance based on the VQA scores~\cite{Antol_2015_ICCV} of the generated answers. 
To align with relevance preference, we employ a direct preference optimization (DPO) algorithm~\cite{rafailov2024direct} to further refine the generative retrieval model.
In the answer generation stage, we input the retrieved documents one by one, and the model obtains the final prediction based on the joint probability of retrieval and answer generation.

We conduct primary experiments on two representative knowledge-based VQA benchmarks, OKVQA and A-OKVQA.
The experimental results show significant improvements of 2.9\%-9.6\% across all metrics compared to strong baselines.
Additionally, we perform knowledge retrieval experiments on three datasets to further validate the performance of the generative knowledge retrievers. 
Our model consistently outperforms other discriminative knowledge retrievers and the improvements become more apparent when applied to large-scale knowledge bases.
This outcome illustrates our model’s capability
to retrieve knowledge from large-scale knowledge sources. The code will be available at \url{https://github.com/xinwei666/ReAuSE}



\section{Related Work}
\textbf{Traditional Visual Question Answering (VQA)} tasks~\cite{johnson2017clevr,mishra2019ocr}, which focus on answering questions related to visual elements (e.g., simple counting, visual attributes), have been extensively studied. 
Several studies~\cite{DBLP:conf/cvpr/MarinoRFM19} have revealed that over 78\% of questions can be answered by people under ten years old, indicating that traditional VQA tasks require little background knowledge to answer a vast majority of questions.


\noindent \textbf{Knowledge-based VQA.} To assess models' capacity to leverage world knowledge instead of relying solely on input data, knowledge-based VQA tasks have emerged, such as OKVQA~\cite{DBLP:conf/cvpr/MarinoRFM19}, and A-OKVQA~\cite{DBLP:conf/eccv/SchwenkKCMM22}. 
OKVQA and A-OKVQA datasets pose challenges in acquiring the necessary knowledge from an outside source and performing reasoning over multi-modal contexts and knowledge. 
Recently, Infoseek~\cite{DBLP:conf/emnlp/ChenHLSCRC23} has been proposed, featuring visual questions about detailed properties of factual knowledge in Wikipedia. The above datasets all highlight the importance of retrieving knowledge from external sources and underscore that current state-of-the-art methods still have significant room for improvement in this task.

Existing approaches have been proposed to incorporate knowledge in two ways to address knowledge-based VQA tasks.
One line of research~\cite{DBLP:conf/emnlp/XenosSPT23,chen2023see,gui2021kat} leverages implicit knowledge from LLMs.
This approach involves converting images into text or directly feeding multi-modal contexts into LLMs (e.g. GPT-3~\cite{brown2020language}, GPT-4V~\cite{achiam2023gpt}, etc.) to generate text that serves as augmented knowledge, but hallucinated information produced by LLMs poses risks to the overall pipeline. 
Another research direction~\cite{DBLP:conf/nips/LinX0X0Y22,DBLP:journals/corr/abs-2403-10037,DBLP:conf/nips/LinCMCB23} aims to retrieve explicit knowledge from structured or unstructured KB.
This approach, known as retrieval augmentation, often uses off-the-shelf tools to generate visual tags and captions, thereby boosting the performance of knowledge retrievers.
Several studies~\cite{DBLP:conf/cvpr/0013PTRWN22,DBLP:conf/cvpr/HuI0WCSSRF23} have tried to combine both ways by simply using the results of LLMs and retrievers but led to limited improvements over baselines.

\noindent \textbf{Knowledge Retrieval.} As a crucial component of retrieval-augmented approaches, knowledge retrievers face challenges in handling multi-modal queries~\cite{DBLP:conf/emnlp/LuoZBB21,luo2023end,shen2023pbsl}. Several methods~\cite{lin2022retrieval,DBLP:conf/emnlp/LinB22,DBLP:conf/cvpr/0013PTRWN22}, which employ separate text-to-text and image-to-text retrievers, struggle to capture cross-modal interactions. 
To bridge this gap, Reviz~\cite{luo2023end} leverages visual-language models to unify the encoding of image and text queries, and FMLR~\cite{DBLP:conf/nips/LinCMCB23} proposes a fine-grained late-interaction framework to fuse cross-modal features at the token level.
PreFLMR~\cite{lin2024preflmr} explores scaling laws for knowledge retrieval based on the FLMR model.
Although these methods achieve improvements over previous approaches, they require training on large-scale datasets containing millions of image-text pairs, which incurs high computational costs.

Recently, some studies~\cite{bevilacqua2022autoregressive,ziems2023large,li2023multiview,li2024survey,DBLP:conf/acl/LongZMZZ24,jain-etal-2024-rag} have introduced generative pipelines in information retrieval tasks, instead of discriminative retrievers.
These methods~\cite{DBLP:conf/nips/Tay00NBM000GSCM22} are based on the assumption that all documents are memorized by generative language models, and the language model directly generates the identifiers of relevant documents based on the query.
While prior research~\cite{li2024generative,long2024generative} has investigated generative retrieval for multi-modal tasks, such methods have demonstrated only marginal gains over traditional methods when applied to general tasks.
Different from them, we are the first work to seamlessly integrate generative retrieval and retrieval-augmented VQA tasks, and use the feedback from the QA module to enhance the retrieval performance, thereby achieving better retrieval and QA results simultaneously.

\begin{figure*}[t]
\centering
\includegraphics[width=1\textwidth]{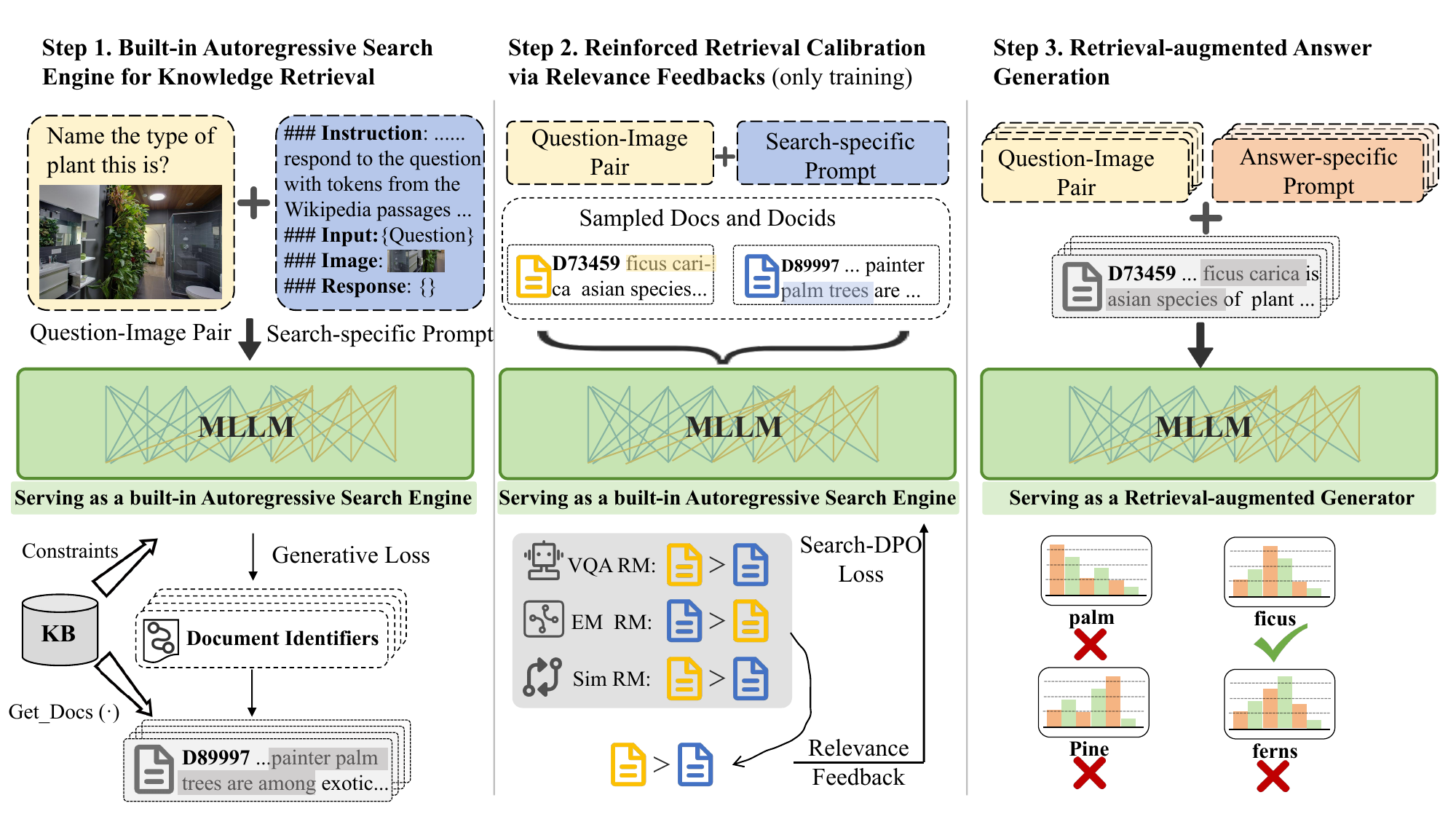} 
\caption{The architecture of ReAuSE. ReAuSE contains three components: Built-in Autoregressive Search Engine for knowledge retrieval, Reinforced Retrieval Calibration via Relevance Feedback to align retrievers with relevance preferences, and Retrieval-Augmented Generation for answer prediction.}
\label{fig2}
\end{figure*}

\section{Methodology}
We introduce ReAuSE, a \textbf{Re}trieval-\textbf{Au}gmented framework utilizing built-in \textbf{Au}toregressive \textbf{S}earch \textbf{E}ngines tailored for knowledge-based VQA tasks.
ReAuSE is designed as
a unified model to facilitate both effective knowledge retrieval and question-answering tasks.

\subsection{Problem Formulation}
Formally, let $\mathcal{D}=\{D_1, ..., D_k\}$ denotes a knowledge base used for the knowledge-based VQA task, $D_i=\{d_1, ..., d_{|D|}\}$ denotes a document with its title and textual contexts, and $R_i=\{r_1, r_2, ..., r_{|R|}\}$ denotes an identifier of the document $D_i$.
Given a multi-modal query $X$, the generative knowledge retrieval can be formulated as a Seq2Seq task, as Eq.~\ref{eq1},
\begin{equation}
  \mathcal{P}(R_i|X) = \prod_{j=1}\mathcal{P}(r_j|\bm {r}_{<j},X,\Theta) . 
  \label{eq1}
\end{equation}
where $\mathcal{P}$ denotes the standard auto-regressive language modeling probability and $\bm \Theta$ are the paramters of our model. During inference, the model employs a constrained strategy to guide the decoder in generating valid identifiers, which maintains a deterministic mapping relationship $\varphi$ between identifier and document, as Eq.~\ref{eq2},
\begin{equation}
  \varphi: R_i \to D_i, {\rm where}\  D_i \in \mathcal{D} .
  \label{eq2}
\end{equation}
Finally, we obtain a subset $\hat{D}=\{D_1, ..., D_{|K|}\}$ from $\mathcal{D}$ to improve answer generation. The overall likelihood of generating the answer $Y$ is given by Eq.~\ref{eq3},
\begin{equation}
  \mathcal{P}(Y|X) =\sum_{D_i \in \hat{D}} \underbrace{\mathcal{P}(R_i|X)}_{retrieval} \cdot \underbrace{\mathcal{P}(Y|X, D_i)}_{generation} . 
  \label{eq3}
\end{equation}


\subsection{Built-in Autoregressive Search Engines}
We introduce a novel autoregressive search engine for knowledge-based VQA tasks to facilitate retrieval from external knowledge bases.
The autoregressive search engine leverages a generative architecture similar to that of common multimodal large language models, instead of discriminative models, enabling its seamless integration and functioning as a built-in module.

Given a multi-modal input $X=\{Q, V\}$, the autoregressive search engine aims to generate the relevant identifier directly in a seq2seq manner as Eq.~\ref{eq1}. 
For example, Fig.~\ref{eq2} shows how our model generates the corresponding identifier for a document related to the ``palm tree" based on the input image and question. 
To achieve such a generative retriever, we mainly elaborate on the three aspects as follows:

\noindent \textbf{Document Identifier.} 
Based on the assumption in ~\cite{pan2024unifying} that large language models are aware of the content within each document, we define any document's identifier as subsequences that appear only in that specific document.
Unlike the one-to-one relationship in DSI~\cite{DBLP:conf/nips/Tay00NBM000GSCM22}, We assign more than one identifier to each document, as long as these identifiers are unique for this document.
Consequently, our model does not require additional memory steps as in existing studies~\cite{DBLP:conf/nips/Tay00NBM000GSCM22,li2024generative} to associate documents with identifiers.

\noindent \textbf{Supervised Fine-tuning} teaches our model to generate relevant identifiers based on the autoregressive probability for each given multi-modal query. 
To sample the most relevant sub-sequences from the given ground-truth document as identifiers, we employ a large language model~\cite{touvron2023llama} as an extractive summarizer, which uses a fixed-length original text to answer a given question.
Later, we filter the obtained set of identifiers and select the identifier containing the most answer keywords as the target identifier.
Note that our model is model-agnostic, allowing it to be applied to any generative multi-modal large language model. 
The generative loss function can be formalized as maximizing the likelihood of the target identifier using the teacher forcing strategy, as Eq.~\ref{eq4}.
\begin{equation}
  \mathcal{L}_{retrieval} = \sum_{j=1} {\rm log} \mathcal{P} (r_j|\bm {r}_{<j},X) . 
  \label{eq4}
\end{equation}
To avoid overfitting and catastrophic forgetting, we freeze all the parameters of the MLLM and adopt the Low-Rank Adaptation (LoRA) method~\cite{hu2021lora} to efficiently fine-tune our model, with only the parameters of LoRA being updated.

\noindent \textbf{Constrained Decoding and FM-Index.} 
A valid identifier is defined as a generated sequence that appears at least once within a document in the knowledge base, ensuring that each generated identifier can be directly linked to a specific document.
To help the model generate valid identifiers during inference, we implement a beam decoding strategy constrained by knowledge bases.

Specifically, we use the previously generated sequence $R_i^{t-1}=\{r_1,...,r_{t-1}\}$ as the prefix condition to search for all matching strings in the knowledge base.
We then extract the subsequent tokens from these strings to form a feasible token set $\mathcal{S}$.
The model's next token, $r_t$, is restricted to selection from $\mathcal{S}$, guaranteeing that all generated sequences exist within the knowledge base.
To support fast substring search, we utilize an FM-Index database~\cite{ferragina2000opportunistic,bevilacqua2022autoregressive} to store the knowledge base. FM-Index is an efficient indexing structure tailored for substring search. The time complexity for obtaining the next allowed token is nearly $\mathcal{O}(V)$, where $V$ is the vocabulary size, independent of the size of the knowledge base.

\subsection{Reinforced Retrieval Calibration via Relevance Feedback}
Despite teaching our model through supervised fine-tuning to generate relevant document identifiers based on user queries, the retrieved documents exhibit varying degrees of relevance. 
Even when documents are provided, the QA model may struggle to provide accurate responses.
Optimally, the generative retriever should retrieve documents that: (1) strongly correlate with the multi-modal query, and (2) minimize extraneous content.
Consequently, it is essential to further improve retrieval performance through feedback from the QA model.

As the first step towards this goal, we sample a set of identifiers $\{R_1,...,R_k\}$ for each $X$ using the generative retriever $\pi_{sft}$ that has been supervised fine-tuned.
Then, we score the collected samples by evaluating their relevance from three aspects:
\begin{itemize}
\item \textbf{Contributions to VQA performance.} 
A document is deemed relevant if a model can produce the correct answer using it. To evaluate this relevance, we employ an MLLM that has not been fine-tuned on downstream data as the reward model, with the VQA score serving as the reward value $v_{vqa} \in [0, 1]$.
\item \textbf{Keyword Hit Count.} If an identifier includes keywords from the answer set, it is likely to be relevant. To quantify this relevance, we employ an exact matching function as the reward function, with matching signals serving as the reward values $v_{hit}\in \{0, 1\}$.
\item \textbf{Semantic Similarity.} Higher semantic similarity between an identifier and a document indicates that the identifier better represents the document's semantics, thereby suggesting a lower presence of irrelevant content within the document. To measure this relevance, we use the BERT model to calculate the cosine similarity between identifiers and documents as the reward values $v_{sim} \in [0, 1]$.
\end{itemize}

The overall reward can be obtained by taking a weighted sum of the scores from different aspects.
Then, we build a triplet $<X, R^{+}, R^{-}>$ for each $X$ by treating the identifiers with the highest/lowest reward as positive/negative samples, respectively. 
Using the triplets reflecting the QA model's preference, the retriever can be further aligned by preference-based reinforcement learning.
As one of the typical methods, direct preference optimization (DPO)~\cite{rafailov2024direct} is widely used for its efficiency and effectiveness.
Therefore, we employ the DPO loss to further optimize our autoregressive knowledge retriever as Eq.~\ref{eq7},
\begin{equation}
  \mathcal{L}_{dpo} = - {\rm log} \sigma\Bigg( \beta {\rm log} \frac{\pi_{\Theta}(R^+|X) \pi_{sft}(R^-|X)}{\pi_{sft}(R^+|X) \pi_{\Theta}(R^-|X)} \Bigg)  . 
  \label{eq7}
\end{equation}
where $\pi_{sft}$ is the original model used as reference, and $\pi_{\Theta}$ is the model being optimized. As before, we only update the parameters of LoRA.

\subsection{Answer Generation}
Utilizing built-in autoregressive knowledge retrievers, we extract the top-K relevant documents from extensive knowledge bases to serve as external knowledge.
For our answer generation model, we employ a model architecture homologous to that of the retrieval module.
As illustrated in Fig. 2, we construct a prompt template, filling the slots with the image, question, and each retrieved document.
The multi-modal contexts are then fed into the model, and the training loss of the answer generation follows that of the generative retrieval model, as Eq.~\ref{eq5},
\begin{equation}
  \mathcal{L}_{gen} = \sum_{j=1} {\rm log} \mathcal{P} (y_j|\bm {y}_{<j},X, D_i) . 
  \label{eq5}
\end{equation}
where $y_j$ denotes the $j-th$ token of the ground-truth answer $Y$. As before, we freeze all the parameters of the MLLM, but introduce another LoRA, and only update the parameters of this new LoRA.
\begin{equation}
\begin{split}
	\hat{Y}, \hat{D} &= \mathop{\rm arg max}\limits_{Y, D_i} \mathcal{P} (Y, D_i|X) \\
	&= \mathop{\rm arg max}\limits_{Y, D_i} \mathcal{P} (Y|X, D_i) \cdot \mathcal{P} (R_i|X).
\label{eq6}	
\end{split}
\end{equation}
During inference, We use the same MLLM and parameters for both the retrieval and answer generation stages, except for the two LoRA adapters. 
After retrieving the relevant document set, we switched to the LoRA adapter for answer generation, and obtain
the final prediction through the joint probability of retrieval and answer generation, as Eq.~\ref{eq6}.

\section{Experiments}
\subsection{Experiment Setup}
\textbf{Datasets and Knowledge Bases.}
We focus on the knowledge-based VQA benchmarks, OKVQA~\cite{DBLP:conf/cvpr/MarinoRFM19} and A-OKVQA~\cite{DBLP:conf/eccv/SchwenkKCMM22}. 
Previous work provided two retrieval corpora, GS112K~\cite{DBLP:conf/emnlp/LuoZBB21} and Wiki21M~\cite{karpukhin2020dense}, for the OKVQA dataset. GS112K contains 112K passages collected through Google Search, while Wiki21M is a subset of Wikipedia, containing 21M Wikipedia entries.
Moreover, we also conduct retrieval experiments on these two corpora and introduce a new information-seeking dataset, InfoSeek~\cite{DBLP:conf/emnlp/ChenHLSCRC23}, to evaluate the model's retrieval performance. Since InfoSeek's KB is not publicly available, we use the KB provided by PreFLMR~\cite{lin2024preflmr} and follow the same experimental setup.

\noindent \textbf{Evaluation Metrics.} 
We strictly follow the settings of the original papers, using the corresponding metrics for each dataset.
For the OKVQA dataset and the ``direct answer" setting of the A-OKVQA dataset, we use the VQA score to evaluate the model's performance.
For the ``multi-choice" setting of the A-OKVQA dataset, we use accuracy for evaluation.
To evaluate the performance of knowledge retrieval, we use the Pseudo-relevance Recall@K (PRR@K)~\cite{DBLP:conf/emnlp/LuoZBB21}, consistent with the baselines.

\noindent \textbf{Baselines.} 
We adopt several baseline methods for comparison, categorized as follows:
1) multi-modal large language models: 
LLaVA-13B~\cite{liu2024visual}, PALm-E-562B ~\cite{chenpali}, and GPT-4V~\cite{achiam2023gpt}.
2) knowledge-enhanced methods via GPT-3/4 APIs: Prophet~\cite{shao2023prompting}, Promptcap~\cite{hu2023promptcap}, FillingGap~\cite{wang2023filling} and REVIVE~\cite{DBLP:conf/nips/LinX0X0Y22}.
3) retrieval-augmented methods:  TwO~\cite{si2023combo}, ReVeaL~\cite{DBLP:conf/cvpr/HuI0WCSSRF23}, GeMKR~\cite{long2024generative}, and FLMR~\cite{DBLP:conf/nips/LinCMCB23}.
For the A-OKVQA dataset, we also add the advanced GPV-2~\cite{DBLP:conf/eccv/SchwenkKCMM22}, SimVQA~\cite{xenos2023simple}, Cola-FT(11B+3B)~\cite{chen2024large} and CKR-VQA~\cite{DBLP:journals/corr/abs-2403-10037} as baselines.

\noindent \textbf{Implementation Details.}
Our framework is model-agnostic. 
In our main experiments, we utilize MiniGPT4-v2-7B as the base model, which employ ViT-L/14 from pre-trained CLIP as the image encoder and LLaMa-v2-7B (Touvron et al. 2023) as the text encoder.
We freeze all parameters of the MLLM, allowing updates only to the LoRA parameters. 
\textbf{We use the same MLLM in the three stages but apply two sets of LoRA parameters to optimize the model respectively: one for retrieval and alignment, and the other for answer generation.} 
Our model is implemented in PyTorch, utilizing version 0.3.0 of the PEFT library, which supports efficient switching between two LoRA adapters during inference. 
Similar to baselines, we use image captions as features to enhance the model's performance.
Each training stage is performed on four NVIDIA A6000 48G GPUs and completed within three hours.

\begin{table}[]
\centering
\small
\begin{tabular}{lcc}
\toprule
Model                       & PRR@K         & Score        \\ \midrule
\multicolumn{3}{c}{\textbf{\textit{Multi-modal Large Language Models}}}               \\
LLaVA-13B~\cite{liu2024visual}                    & -                  & 61.9             \\
Minigpt4-v2-7B~\cite{chen2023minigpt}                & -                  & 57.8             \\
Minigpt4-v2-7B (FT)~\cite{chen2023minigpt}                & -                  & 61.9             \\
PaLM-E-562B~\cite{driess2023palm}                 & -                  & 66.1             \\
GPT-4V~\cite{achiam2023gpt}                      & -                  & 64.3             \\ \midrule
\multicolumn{3}{c}{\textbf{\textit{Knowledge-enhanced Methods via GPT-3/4v APIs}}} \\
ReVIVE~\cite{DBLP:conf/nips/LinX0X0Y22}                      & -                  & 58.0             \\
Prophet~\cite{shao2023prompting}                     & -                  & 61.1             \\
Promptcap~\cite{hu2023promptcap}                   & -                  & 60.4             \\
FillingGap~\cite{wang2023filling}                  & -                  & 61.3             \\
MM-Reasoner~\cite{khademi2023mm}                 & -                  & 60.8             \\ \hline
\multicolumn{3}{c}{\textbf{\textit{Retrieval-augmented Generation Methods}}}                  \\
TRiG~\cite{DBLP:conf/cvpr/0013PTRWN22}                        & 45.8                  & 50.5             \\
RA-VQA~\cite{DBLP:conf/emnlp/LinB22}                      & 82.8               & 54.5             \\
TwO~\cite{si2023combo}                         & -                  & 56.7             \\
ReVeaL~\cite{DBLP:conf/cvpr/HuI0WCSSRF23}                      & -                  & 59.1             \\
FLMR~\cite{DBLP:conf/nips/LinCMCB23}                        & 89.3               & 62.1             \\
FLMR~\cite{DBLP:conf/nips/LinCMCB23} $\ast$                        & 88.3               & 62.7             \\
KSVQA~\cite{hao2024boter}                       & -                  & 62.8             \\
GeMKR~\cite{long2024generative} $\ast$                       & 78.6               & 61.8             \\ \midrule
\textbf{ReAuSE (Ours)}          & \textbf{92.6}               & \textbf{65.7}             \\ \bottomrule
\end{tabular}
\caption{Performance on the OKVQA benchmark. PPR@K applies only to RAG baselines; ``-" denotes inapplicability or unavailable results. ``$\ast$" indicates the results we reproduced using the official code and the same answer generator as our model.\protect\footnotemark}
\label{tab_mr_okvqa}
\end{table}

\footnotetext{We first use officially released checkpoints to obtain retrieved documents and then feed these documents into the answer generator (fine-tuned Minigpt4-v2) to acquire the corresponding answers.}

\subsection{Main Results}
We compare our ReAuSE with the aforementioned baselines for knowledge-based VQA tasks in Tab.~\ref{tab_mr_okvqa} and Tab.~\ref{tab_mr_aokvqa}. The experimental results illustrate
that ReAuSE achieves significant improvements over the competitive baselines on the challenging OKVQA and A-OKVQA datasets.

From Tab.~\ref{tab_mr_okvqa}, we can observe that ReAuSE outperforms the competitive baseline FLMR on both retrieval and VQA metrics, which consistently demonstrates the effectiveness of our method in integrating both knowledge retrieval and answer generation into a unified multi-modal large language model framework. 
ReAuSE achieves an advanced VQA score on OKVQA when compared to models with similar parameter scales, surpassing the previous best retrieval-augmented method by more than 2.9\% and outperforming methods that use LLM-APIs for knowledge enhancement by 4.6\%. Moreover, our method exceeds GPT-4V by 1.45\% in VQA score. Even compared with the closed-source PALM-E-562B, which is over 80 times larger than ours, our method is only 0.5\% behind.

The OKVQA benchmark poses a challenging issue of 
retrieving relevant knowledge from extensive knowledge bases or directly generating useful information about multi-modal contexts. 
Despite using GPT-3 or GPT-4V to acquire knowledge or directly adopting GPT-3 as the backbone, MM-Reasoner and FillingGap fail to achieve obvious improvements compared to retrieval-augmented methods. In contrast, retrieval-augmented methods, such as FLMR and KSVQA, achieve better VQA performance by incorporating manually designed feature engineering and integrating multiple retrievers and selectors.

\begin{table}[]
\small
\centering
\begin{tabular}{lcccc}
\toprule
\setlength{\tabcolsep}{5pt}
\multirow{2}{*}{Models} & \multicolumn{2}{c}{Multi-Choice} & \multicolumn{2}{c}{Direct-Answer} \\
                        & val              & test             & val             & test            \\ \hline
LLaVA-1.5-7B            & 77.1             & 74.5             & 63.7            & 58.6            \\
InstructBLIP-7B(FT)         & 73.0             & 71.1             & 62.4            & 58.7            \\
Minigpt4-v2-7B(FT)        & -                & -                & 61.3               & -               \\
GPV-2                   & 60.3             & 53.7             & 48.6            & 40.7            \\
PromptCap               & 73.2             & 73.1             & 56.3            & 59.6            \\
Prophet                 & 76.4             & 73.6             & 58.2            & 55.7            \\
FillingGap              & -                & -                & 59.8            & -               \\
SimVQA                  & -                & -                & 58.6            & 57.5            \\
REVEAL                  & -                & -                & 52.2            & -               \\
Cola-FT                 & 78.1             & 76.7             & -               & -               \\
CKR-VQA                 & 76.2             & 75.4             & 58.1            & 60.1            \\ \midrule
\textbf{ReAuSE (Ours)}      & \textbf{85.0}                & \textbf{80.3}            & \textbf{67.7}               & \textbf{65.8}           \\ \bottomrule
\end{tabular}
\caption{Performance on the A-OK-VQA benchmark.}
\label{tab_mr_aokvqa}
\end{table}

From Tab.~\ref{tab_mr_aokvqa}, 
ReAuSE demonstrates more significant performance improvements on A-OKVQA, with accuracy and VQA scores increasing by 4.9\% to 9.6\% compared to baselines of similar parameter scales\footnote{See the submission at: \url{https://leaderboard.allenai.org/a-okvqa/submission/cqp56m03c8g0k0quidj0}}.
Our approach demonstrates consistent improvements, which can be attributed to two key factors. First, we leverage large language models as virtual knowledge bases by replacing traditional discriminative pipelines with generative retrievers. Second, we implement reinforced retrieval calibration to align the search engine with the answer generator, enabling the retriever to incorporate relevance feedback for refinement, thereby yielding more relevant results.
In the following sections, we will examine the performance of the autoregressive search engine and analyze the impact of search results on the answer generation process.

\begin{table}[t]
\small
\centering
\begin{tabular}{lcc}
\toprule
Ablation Setting                        & PRRecall@5 & Score \\ \midrule
\textbf{Full Model (Ours)}                            & \textbf{92.6}          & \textbf{65.7}      \\ \midrule
\textit{w/o} Search Engines          & -          & 61.9         \\
\textit{w/o} Fine-tuning Search Engine         & 33.1          & 61.6         \\
\textit{w/o} Constrained Decoding              & -          & 63.2         \\
\hline
\textit{w/o} Retrieval Calibration & 88.7          & 62.5         \\
\textit{w/o} VQA Reward Model                 & 91.0          & 63.3         \\
\textit{w/o} EM Reward Func.                  & 89.9          & 64.5         \\
\textit{w/o} Sim. Reward Model          & 91.7          & 65.3         \\ \bottomrule
\end{tabular}
\caption{Ablation Studies. \textit{w/o} denotes ``without".}
\label{tab_as}
\end{table}

\subsection{Ablation Study}
We conduct a series of ablation studies by gradually removing each module of our framework and the corresponding
results are presented in Tab.~\ref{tab_as}.

To evaluate the impact of retrieval augmentation, we first remove the built-in autoregressive search engine, using the MLLM as an answer generator without access to external knowledge. This operation results in a 3.8\% decrease in the VQA score, indicating that external knowledge retrieval is crucial for knowledge-based VQA tasks.
Next, if we do not supervised fine-tune the MLLMs, it cannot effectively serve as a generative search engine to retrieve knowledge from the KB.
Moreover, we disable the constrained decoding strategy, allowing the MLLM to generate image-related knowledge without restrictions. However, since this freely generated content cannot be linked to the document in the KB, it is used directly as external knowledge to support the answer generation process.
This approach leads to a 2.5\% decrease in the VQA score, likely due to the MLLM producing erroneous or hallucinated information, which results in inaccurate outputs from the answer generator.

To evaluate the effectiveness of the Reinforced Retrieval Calibration (RRC) module, we employ the generative search engine after supervised fine-tuning but remove the reinforced calibration module. We observe a 3.9\% decrease in retrieval performance, which is slightly below that of the strongest baseline, FLMR. This suggests that the autoregressive retriever can be further optimized through the RRC module by leveraging relevance feedback from reward models. 
Furthermore, we disable each reward model to assess its effectiveness. We find that the VQA reward model enables the generative retriever to retrieve documents that align with the answer generator's preferences, thereby improving VQA performance. Conversely, the EM reward model ensures that the generated identifiers include answer keywords, leading to enhanced retrieval performance.

\begin{table*}[!t]
\small
\centering
\begin{tabular}{ccccccc}
\toprule
\multirow{2}{*}{\#} & \multirow{2}{*}{Retrievers} & \multicolumn{2}{c}{\begin{tabular}[c]{@{}c@{}}OKVQA- GS112K\end{tabular}} & \multicolumn{2}{c}{\begin{tabular}[c]{@{}c@{}}OKVQA- WK21M\end{tabular}} & InfoSeek-100K \\
                    &                             & PRRecall@5                          & PRRecall@10                          & PRRecall@5                          & PRRecall@10                         & PRRecall@5    \\ \midrule
1                   & DPR~\cite{karpukhin2020dense}                       & 83.4                               & 90.3                                & 66.9                               & 76.4                               & -             \\
2                   & RA-VQA~\cite{DBLP:conf/emnlp/LinB22}                      & 82.8                               & 89.0                                & -                                   & -                                   & -             \\
3                   & ReViz-ICT~\cite{luo2023end}                   & 73.4                                & 83.2                                 & 61.9                                & 72.6                                & -             \\
4                   & GeMKR~\cite{long2024generative}                       & 78.6                                & 86.2                                 & 70.8                                & 79.1                                & 48.9             \\
5                   & FLMR~\cite{DBLP:conf/nips/LinCMCB23}                        & 89.3                               & 94.0                                & 68.1                               & 78.0                               & 47.1         \\
6                   & Pre-FLMR~\cite{lin2024preflmr}                    & -                                   & -                                    & 68.6                                & -                                   & 57.8         \\ \midrule
7                   & \textbf{ReAuSE (Ours)}                      & \textbf{92.6}                                   & \textbf{95.8}                                    & \textbf{88.0}                                   & \textbf{91.3}                                   & \textbf{59.5}             \\ \bottomrule
\end{tabular}
\caption{Retrieval Performance on Three Retrieval Corpora.}
\label{rs_retr}
\end{table*}

\subsection{Effects of Retrieval Performance}
To assess our model's capability in retrieving knowledge from large-scale knowledge bases, we conduct experiments on the OKVQA dataset using two retrieval corpora: Google Search (GS112K)~\cite{DBLP:conf/emnlp/LuoZBB21} and Wikipedia (Wiki21M)~\cite{karpukhin2020dense}, with knowledge bases ranging in size from 112K to 21M documents. 
Additionally, we introduce a new dataset, Infoseek~\cite{DBLP:conf/emnlp/ChenHLSCRC23}, consisting of 100K documents, which poses challenges for visual entity retrieval.
As shown in Tab.~\ref{rs_retr},
our proposed approach consistently outperforms the leading
state-of-the-art baselines FLMR and Pre-FLMR across all evaluated metrics.
Specifically, our model outperforms FLMR by 3.3\% in PRRecall@5 on the GS112K corpus. 
This improvement explains why our answer generation model surpasses the FLMR model by 3.6\% in the VQA score, as shown in Tab.~\ref{tab_mr_okvqa}.
Moreover, our method outperforms FLMR by 10.6\% on the Infoseek dataset and surpasses PreFLMR by 1.7\%, indicating the effectiveness of ReAuSE in handling visual entity retrieval tasks.

We observe a significant performance drop of over 20\% in the FLMR model when applied to the Wiki21M corpus, while our model exhibits only a 4.6\% decrease. This indicates that our model demonstrates stronger generalization capabilities for retrieving from large-scale corpora.
This can be attributed to the advantages of generative search engines, which generate document identifiers through token-level search rather than relying on one-to-one matches at the document level (i.e., document-level search).
Although the number of documents increases, the size of the token set (i.e., vocabulary) does not expand proportionally. Consequently, generative search engines are less affected by the scale of the knowledge base, whereas the performance of discriminative methods degrades as the corpus size increases.


\begin{figure}[t]
\centering
\includegraphics[width=1\columnwidth]{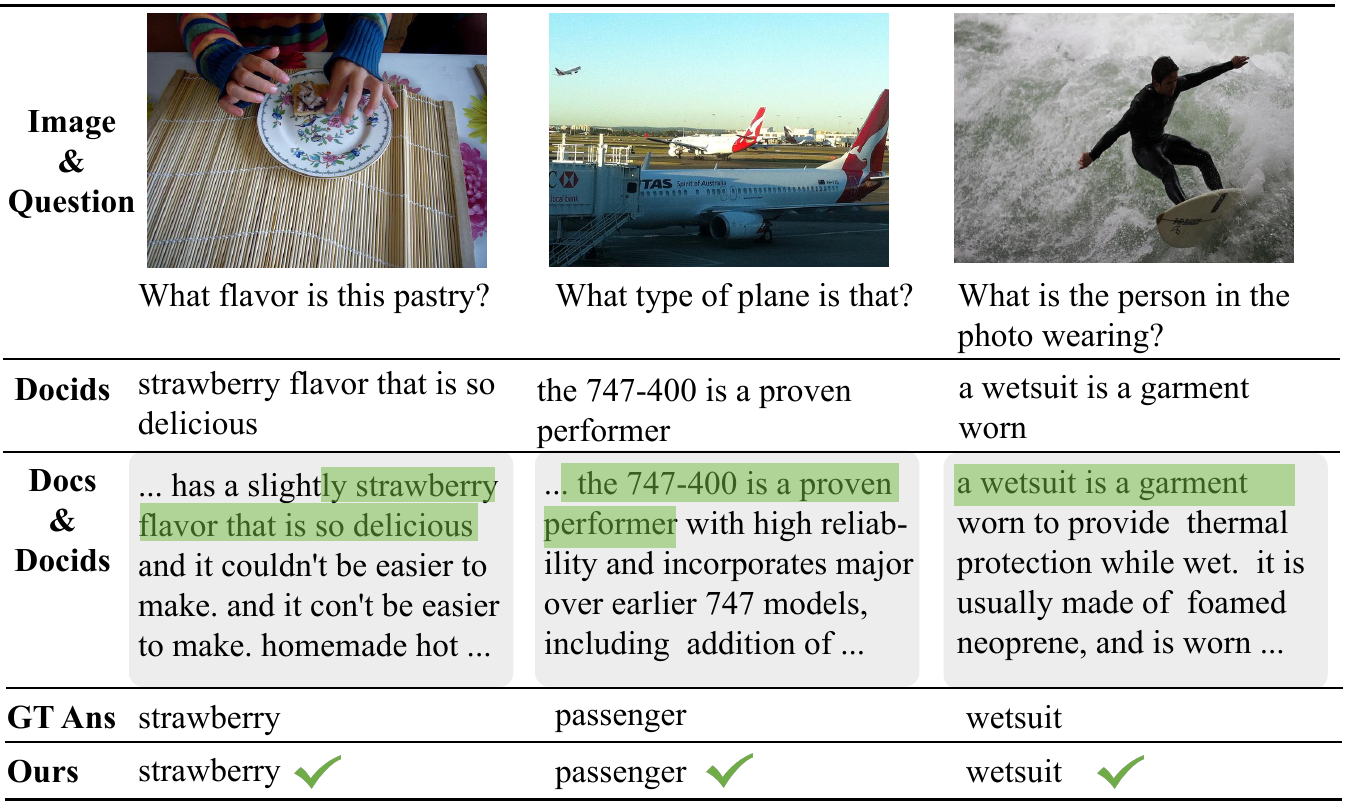} 
\caption{Case Studies. The highlighted text represents the document identifier generated by our model.}
\label{fig3}
\end{figure}



\subsection{Efficiency}
\begin{table}[]
\small
\centering
\begin{tabular}{cccc}
\toprule
  \textbf{Models}  & TopK & GS112K  & Wiki21M \\ \midrule
DPR & 5 & 370.4ms & 518.4ms \\
FLMR & 5  & 758.4ms & - \\
ReAuSE (Ours) & 5 & 751.3ms & 962.1ms \\ 
ReAuSE (Ours) & 10 & 1023.3ms & 1273.2.1ms \\ 
\bottomrule
\end{tabular}
\caption{The Retrieval Time in the Inference Stage.}
\label{tab:ie}
\end{table}
ReAuSE requires fewer resources than other retrieval-augmented baselines.
ReAuSE unifies three stages into a single MLLM, whereas other baselines require an MLLM for generation and at least one traditional retriever for retrieval. 
ReAuSE uses LoRA fine-tuning, requiring 9K (OKVQA) to 17K (A-OKVQA) training data to train 0.49\% of the parameters, with both SFT and RRC completing within 3 hours across 4 GPUs. In contrast, traditional retrievers such as PreFLMR and ReViz-ICT necessitate additional millions of data for full-scale fine-tuning.

We record the inference time of ReAuSE, FLMR, and the most efficient baseline, DPR to provide a qualitative result. As shown in Tab.~\ref{tab:ie}, ReAuSE has comparable efficiency to traditional retrievers. 
ReAuSE generates top-K document identifiers with $l$ tokens for each query through a $l$-steps decoding ($l=10$). In contrast, for each query, traditional retrievers need to calculate the similarity with many documents. 
When TopK=5, our model is only 440 milliseconds slower than DPR. Considering the nearly 20\% performance improvement, we argue that such latency is acceptable.




\subsection{Case Study}
As illustrated in Fig.~\ref{fig3}, ReAuSE accurately generates the correct answers for all three samples. 
ReAuSE directly generates document identifiers associated with the image-text pairs using its built-in search engine. 
Each document identifier is a sequence of tokens representing a document, and it can be linked to a corresponding document that potentially contains information to answer the given question.
What's more, we observe that all generated document identifiers contain answer keywords, suggesting that generated document identifiers are highly relevant to the question.

\section{Conclusion}
In this paper, we introduce ReAuSE, a novel KBVQA approach by integrating knowledge retrieval and generation within a unified generative multi-modal large language model (MLLM) framework. Extensive experimental results have shown that ReAuSE consistently outperforms existing methods, achieving significant improvements across various evaluation metrics on two benchmarks. 
Future work will focus on extending the application of ReAuSE to domains such as biomedicine and education~\cite{gao2021rcd}.
\section{Acknowledgments}
This work was supported by the National Key Research and Development Program of China (2022ZD0160603), the NSFC (No. 62406161), CPSF (No. 2023M741950), and the Postdoctoral Fellowship Program of CPSF (No. GZB20230347).
\bibliography{aaai25}
\appendix
\clearpage

\end{document}